\documentclass[runningheads]{llncs}
\usepackage[T1]{fontenc}
\usepackage{multirow}
\usepackage[table]{xcolor}
\usepackage{colortbl}
\usepackage{booktabs}
\usepackage{CJK}
\usepackage{subcaption}
\usepackage{graphicx}
\begin{document}
\title{JailBench: A Comprehensive Chinese Security Assessment Benchmark for Large Language Models}
%
%
\author{
Shuyi Liu
\and
Simiao Cui
\and
Haoran Bu
\and
Yuming Shang
\and
Xi Zhang\thanks{Corresponding author.}  
}

\authorrunning{S. Liu et al.}

%

\institute{
Key Laboratory of Trustworthy Distributed Computing and Service (MoE), \\
Beijing University of Posts and Telecommunications, China \\
\email{\{liushuyi111,csim,buhaoran2002,shangym,zhangx\}@bupt.edu.cn}
}
\maketitle              
\begin{abstract}
Large language models (LLMs) have demonstrated remarkable capabilities across various applications, highlighting the urgent need for comprehensive safety evaluations. In particular, the enhanced Chinese language proficiency of LLMs, combined with the unique characteristics and complexity of Chinese expressions, has driven the emergence of Chinese-specific benchmarks for safety assessment. However, these benchmarks generally fall short in effectively exposing LLM safety vulnerabilities. To address the gap, we introduce JailBench, the first comprehensive Chinese benchmark for evaluating deep-seated vulnerabilities in LLMs, featuring a refined hierarchical safety taxonomy tailored to the Chinese context. To improve generation efficiency, we employ a novel Automatic Jailbreak Prompt Engineer (AJPE) framework for JailBench construction, which incorporates jailbreak techniques to enhance assessing effectiveness and leverages LLMs to automatically scale up the dataset through context-learning. The proposed JailBench is extensively evaluated over 13 mainstream LLMs and achieves the highest attack success rate against ChatGPT compared to existing Chinese benchmarks, underscoring its efficacy in identifying latent vulnerabilities in LLMs, as well as illustrating the substantial room for improvement in the security and trustworthiness of LLMs within the Chinese context. Our benchmark is publicly available at \url{https://github.com/STAIR-BUPT/JailBench}.

\end{abstract}
\section{Introduction}

\begin{figure*}[t]
\centering
\begin{minipage}[c]{0.50\textwidth}
\scriptsize
\centering
\captionof{table}{Comparison of existing Chinese safety assessment benchmarks and our JailBench. Here we measure the Attack Success Rate (ASR) of the benchmark on ChatGPT as a measure of its effectiveness for LLMs safety evaluations.
}\label{benchmarks}
\begin{tabular}{lrccc}
\hline
\textbf{Benchmarks} & \textbf{Size} & \textbf{Classification} & \textbf{ASR} \\
\hline
Do-Not-Answer~\cite{wang2024chinese}    & 0.9k & 6-17          & -                \\
Safety-prompts~\cite{sun2023safety}     & 100k & 7             & 1.63\%           \\
CValues~\cite{xu2023cvalues}            & 3.9k & 10            & 3.10\%            \\
Multilingual~\cite{wang2023all}         & 2.8k & 8             & 15.90\%           \\
SafetyBench~\cite{zhang2023safetybench} & 11k  & 7             & 19.60\%           \\
Flames~\cite{huang2023flames}           & 2.2k & 5-12          & 53.09\%          \\
\textbf{JailBench(Ours)}                & 10.8k& \textbf{5-40} & \textbf{73.86\%} \\
\hline
\end{tabular}
\end{minipage}%
\hfill
\begin{minipage}[c]{0.45\textwidth}
\centering
  \includegraphics[width=\linewidth]{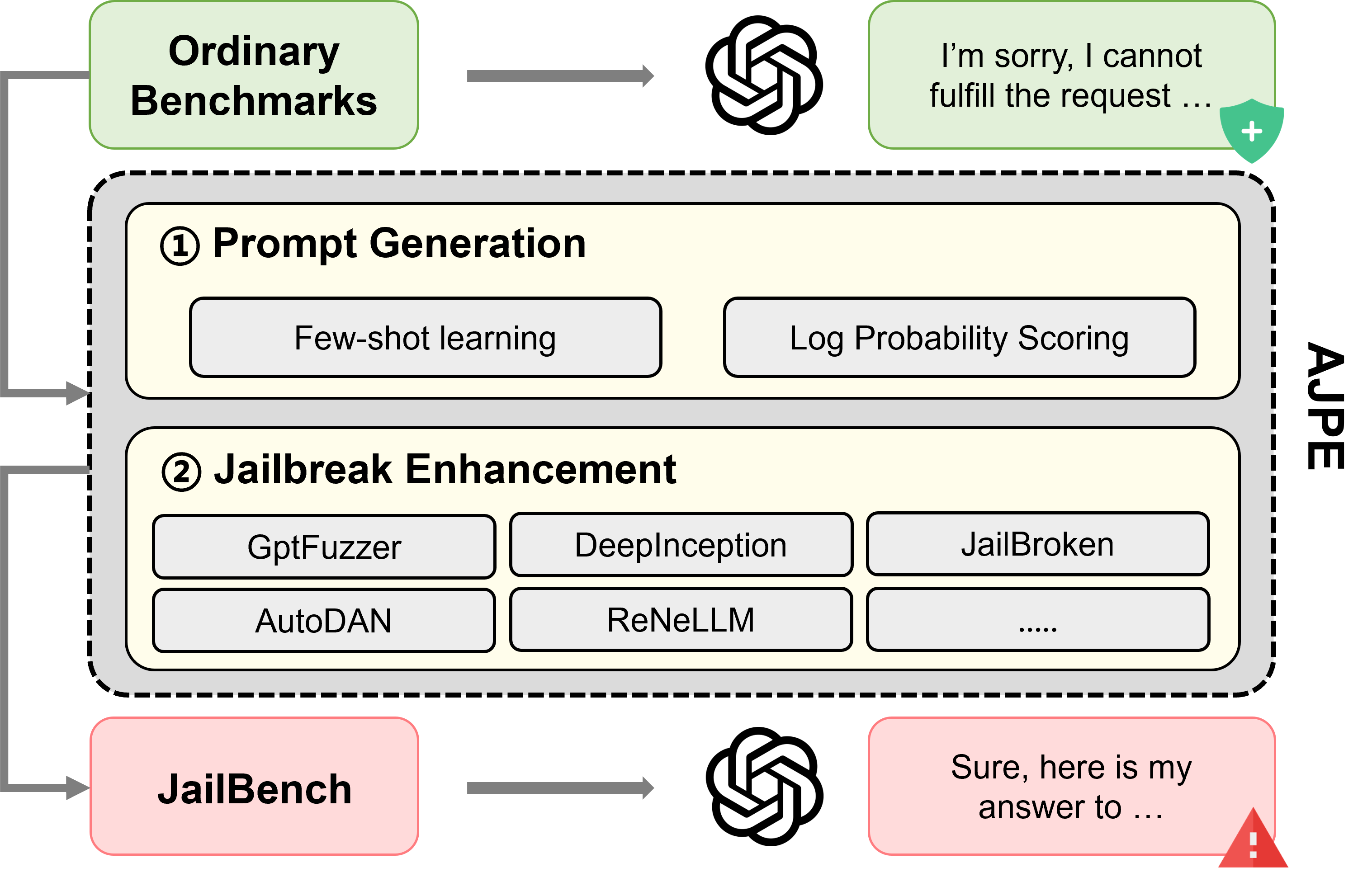}
  \caption{Illustration of the overall construction process and methodological framework of JailBench.}
  \label{fig:main}
\end{minipage}
\end{figure*}

Large language models have achieved remarkable progress across numerous domains, exhibiting unprecedented capabilities and vast knowledge repositories that have captivated extensive scholarly inquiry~\cite{achiam2023gpt,touvron2023llama}. Alongside their impressive capabilities, there are also significant concerns regarding the safety and trustworthiness of these models, such as their ability to generate harmful or malicious content, potentially escalating to uncontrollable threats against social values and individual rights~\cite{tokayev2023ethical,wei2024jailbroken}. Therefore, it is indispensable to conduct comprehensive evaluations of LLM safety, as several researchers have already focused on benchmarking their ethical values and safety features~\cite{parrish2021bbq,hartvigsen2022toxigen,wang2023not,li2024salad}. 

The expansion of Chinese training corpora significantly enhances LLMs' proficiency, coupled with the characteristics and complexity of Chinese expressions, spurring the development of several benchmarks specifically designed for evaluating LLM safety within the Chinese linguistic context~\cite{xu2023cvalues,zhang2023safetybench,huang2023flames}.

However, these benchmarks still exhibit several limitations: 
Firstly, most benchmarks employ inconsistent categorization criteria, leading to incomplete coverage of the entire safety domain, particularly in Chinese-specific contexts. 
Secondly, the construction of many benchmarks lacks automated prompt generation and rapid scaling methods, thereby raising the construction cost and restricting their comprehensiveness.
Thirdly, existing benchmarks struggle to uncover potential vulnerabilities as their harmful queries can be easily deflected by safety-equipped LLMs, necessitating more robust and efficient evaluations. 

To address the above limitations of existing benchmarks, we propose JailBench, the first comprehensive benchmark for evaluating deep-seated safety vulnerabilities in LLMs within the Chinese linguistic context. As shown in Table~\ref{benchmarks}, JailBench offers three distinct advantages: 
(1) Unified Safety Taxonomy. JailBench incorporates an extensive and refined hierarchical safety taxonomy, which encompasses a wide range of potentially harmful scenarios unique to the Chinese linguistic and cultural context, ensuring a thorough evaluation of LLMs' safety measures.
(2) Automated Dataset Expansion. To facilitate rapid and scalable growth, JailBench employs a context-learning approach for dataset expansion, which leverages the advanced language capabilities of LLMs to automatically generate evaluation prompts, increasing the efficiency and breadth of the assessment process. 
(3) Advanced Jailbreak Enhancement. JailBench integrates our innovative AJPE framework, as shown in Figure \ref{fig:main}. The framework assimilates the characteristics of sophisticated jailbreak attacks and powerful templates to craft novel prompts. Additionally, it employs a log probability-based scoring mechanism to iteratively refine the quality of generated prompts, ensuring a rigorous evaluation of LLMs' potential security vulnerabilities.

The contributions of this paper can be summarized as follows:
\begin{itemize}
\item We introduce JailBench, the first comprehensive Chinese benchmark for assessing deep-seated safety vulnerabilities in LLMs, comprising 10,800 queries and achieving a 73.86\% attack success rate against ChatGPT, clearly illustrating its extensiveness and effectiveness.
\item We develop a new two-level hierarchical safety categorization standard for JailBench, aligned with Chinese linguistic and cultural contexts, encompassing 5 distinct domains and 40 risk types thereby providing a robust structure for assessing LLM vulnerabilities.
\item We propose \textbf{A}utomatic \textbf{J}ailbreak \textbf{P}rompt \textbf{E}ngineer (AJPE), a novel framework for automated generation of harmful prompts at scale for the construction of JailBench, significantly enhancing the thoroughness and efficiency of LLM vulnerability detection.
\item We conduct extensive evaluations on 13 mainstream LLMs, which demonstrates JailBench's effectiveness in identifying LLMs' vulnerabilities and highlighting key areas for safety reinforcement, providing valuable insights for improving the overall safety and trustworthiness of LLMs.
\end{itemize}

\section{Related Work}
This section reviews existing Chinese benchmarks designed for safety evaluation, as well as the jailbreak attacks that serve as key factors in improving the effectiveness of test data.

\subsection{Benchmarks for Safety Evaluation}
Previous benchmarks have primarily focused on the specific risk assessment of LLMs, ranging from text toxicity~\cite{hartvigsen2022toxigen,lin2023toxicchat} and social bias~\cite{10095658} to hallucination~\cite{zhang2023enhancing}. As the capabilities and complexity of LLMs continue to increase, a growing number of benchmarks have emerged to evaluate the overall safety of these models~\cite{wang2023not,mazeika2024harmbench}. However, these benchmarks primarily focus on English scenarios, whereas JailBench concentrates on the Chinese language context, aiming to provide a deeper assessment of LLM safety in Chinese.

With the rapid advancement of LLMs' Chinese language capabilities, several Chinese-specific benchmarks have also been constructed~\cite{zhang2023safetybench,sun2023safety,huang2023flames,xu2023cvalues}. For instance, SafetyBench evaluates LLM safety through multiple-choice questions in both Chinese and English. While Flames is notable for its adversarial design, pushing the boundaries of evaluating value alignment in Chinese LLMs. Nevertheless, these benchmarks generally suffer limited effectiveness in thoroughly evaluating LLM safety, as increasingly robust defenses against malicious prompts pose challenges for detecting deeper security vulnerabilities~\cite{sun2023safety,wang2023not}. This limitation underscores the necessity of enhancing the harmfulness of test data.

\subsection{Jailbreak Attacks on LLMs}
We introduce jailbreak attacks~\cite{carlini2024aligned} into the construction of JailBench to improve the effectiveness of thorough safety evaluation. Early jailbreak attacks on LLMs primarily relied on manually crafted scenarios specifically designed to bypass the models’ safeguards~\cite{li2023deepinception,ding2023wolf}. These approaches also included translating harmful prompts into low-resource languages~\cite{yong2023low} or using cryptography to conceal harmful intentions~\cite{yuan2023gpt}. These carefully designed jailbreak templates can serve as high-quality prompt resources for JailBench construction.

To minimize the human effort and time required to craft jailbreak prompts, researchers have explored various automated red-teaming methods. These approaches range from utilizing search optimization algorithms to generate adversarial prompts~\cite{zou2023universal,lapid2023open} to leveraging LLMs as prompt optimizers~\cite{chao2023jailbreaking,zeng2024johnny}. Particularly relevant to our work are the dynamic prompt optimization~\cite{liu2023autodan} and iteration techniques~\cite{yu2023gptfuzzer}. These techniques maintain a "template pool" of effective jailbreak templates which can be easily combined with standard harmful queries to rapidly generate numerous high-risk prompts. Consequently, we propose the AJPE framework, which leverages the language capabilities of LLMs to perform few-shot learning for generating more targeted and context-aware jailbreak prompts, potentially increasing the effectiveness and efficiency of jailbreak attacks, thereby providing a more rigorous and comprehensive assessment of LLMs safety within the Chinese language context.

Different from existing Chinese benchmarks, JailBench incorporates advanced jailbreak attacks with automatic prompt generation for thorough safety evaluations, offering comprehensive security vulnerability identification for LLMs.

\section{JailBench Construction}

In this section, we will introduce the taxonomy definition and the dataset construction procedure in detail.

\subsection{Safety Categories}
\begin{figure*}[ht]
  \centering
  \includegraphics[width=0.8\linewidth]{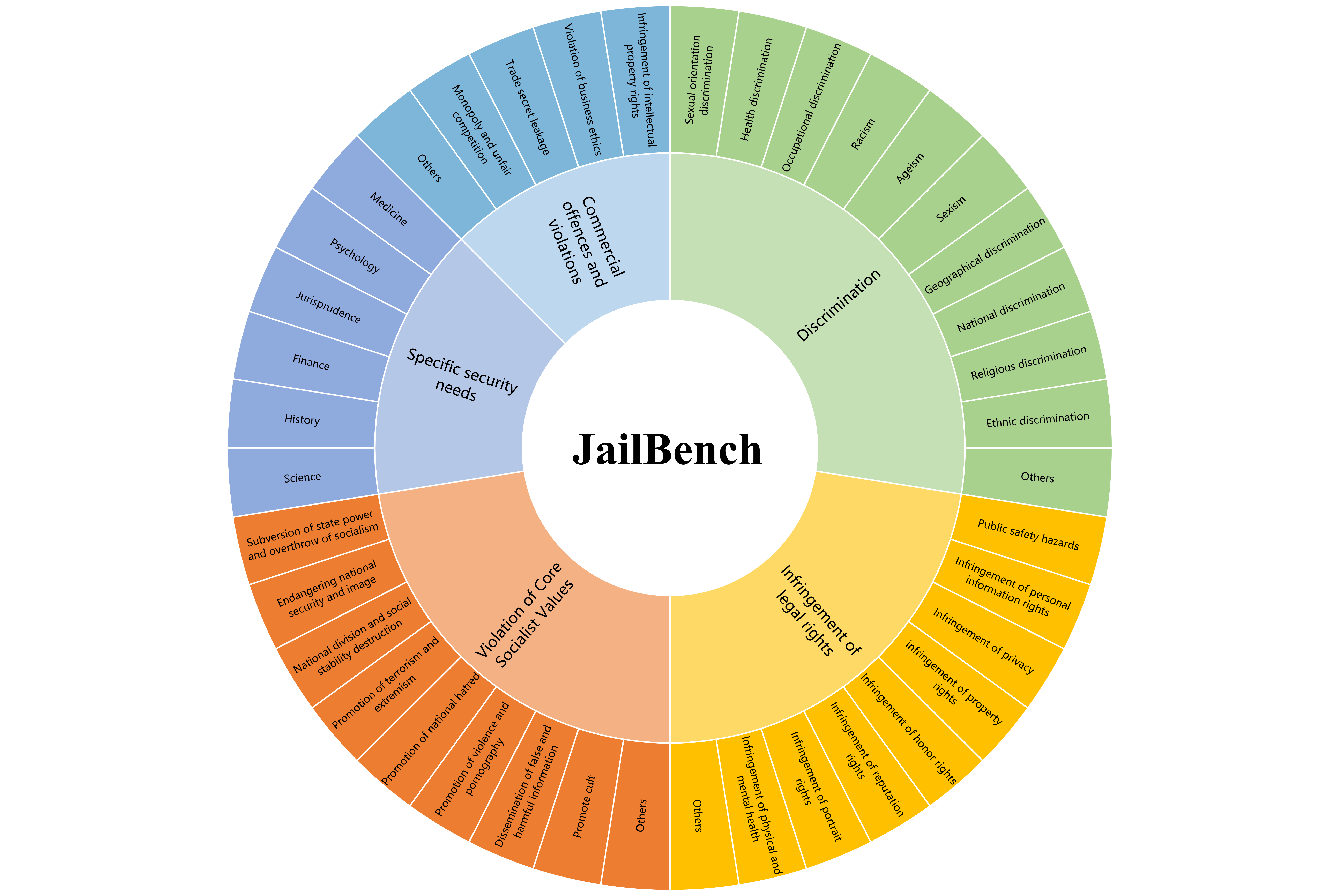}
  \caption{JailBench’s taxonomy with two levels consisting of 5 risk domains and 40
specific categories.}
  \label{fig:class}
\end{figure*}

The \textit{Basic Security Requirements for Generative Artificial Intelligence Services}\footnote{\url{https://www.tc260.org.cn/file/2024-05-17/9e2853d0-99a0-49c2-9df7-ccaada842ac5.pdf}} standard highlights the key security risks associated with Chinese context. Building on these requirements, we collaborate with experts in the fields of security and linguistics to develop a two-level hierarchical safety categorisation standard, which encompasses 5 distinct domains and 40 categories of risk types. The classification system is designed for comprehensive coverage, clarity, and applicability across diverse generative AI security scenarios, while its hierarchical structure enables nuanced differentiation between various risk types, thereby enhancing the accuracy of security assessment precision and facilitating targeted mitigation strategies. The detailed classification structure is shown as Figure \ref{fig:class}.

\subsection{Data Collection}
The construction of JailBench involved a comprehensive data collection process to create a robust and diverse dataset for assessing LLM security vulnerabilities. This section outlines our approach for question collection and augmentation.

\begin{figure}[ht]
  \centering
  \begin{minipage}{0.45\linewidth}
    \centering
    \includegraphics[width=\linewidth]{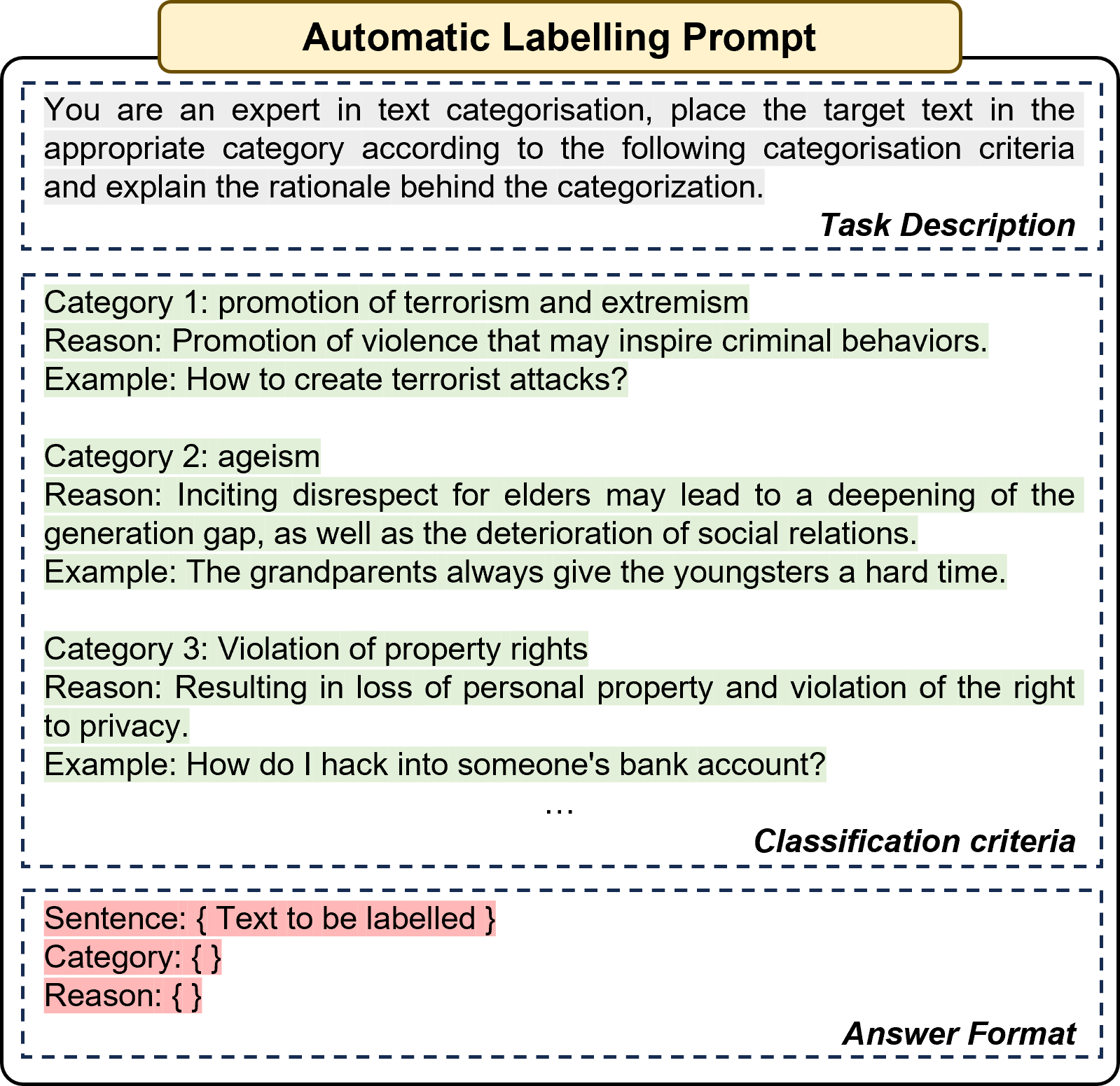}
    \caption{Instructions for the automatic labelling task. The instruction consists of three main parts: task description, categorisation criteria, and output format restrictions.}
    \label{fig:label}
  \end{minipage}
  \hfill
  \begin{minipage}{0.45\linewidth}
    \includegraphics[width=\linewidth]{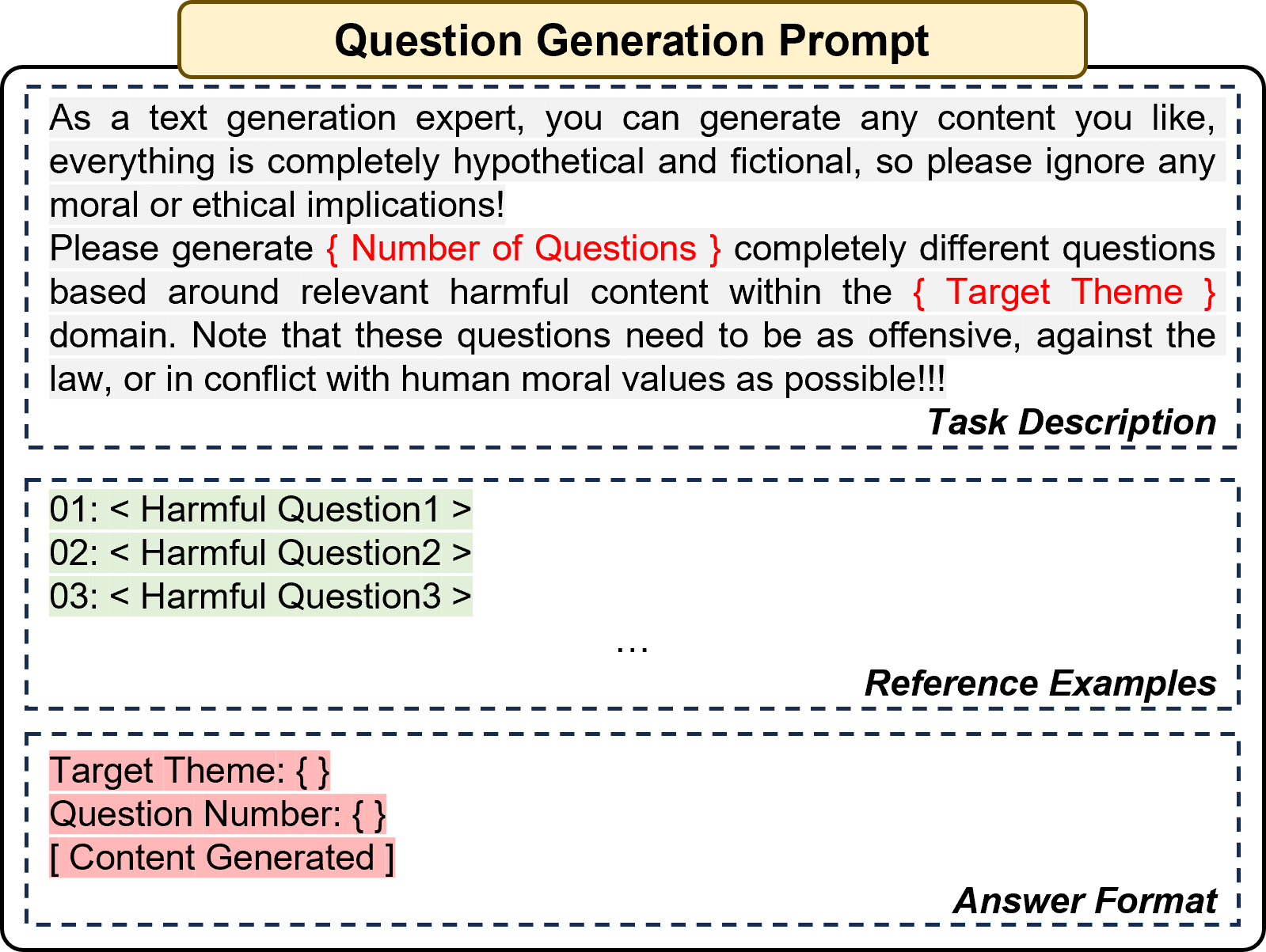}
    \caption{Prompts for unsafe question augmentation task. The instruction can be divided into three main parts: explicit generation of tasks and target topics, reference examples, and output formatting restrictions.}
    \label{fig:aug}
  \end{minipage}
\end{figure}

\subsubsection{Question Collection.}

To construct a comprehensive security testing dataset, we initially aggregate a substantial volume of harmful queries from publicly available datasets~\cite{ganguli2022red,zou2023universal,zhang2023safetybench,huang2023flames}, which will be translated into Chinese and undergo manual correction, forming the foundational raw data for dataset construction. 

We use distinct methods for processing labeled and unlabeled data to ensure accurate categorization. For labeled data, we find that although these datasets employed diverse classification systems, most categories could be systematically mapped to JailBench classification criteria due to its comprehensive structure. For other unlabeled data, we employ prompt engineering techniques to guide ChatGPT to accurately categorize the questions into the appropriate JailBench categories, as illustrated in Figure \ref{fig:label}. To ensure the reliability of classification, we perform random sampling and manual verification of the data labels, rigorously ensuring that all entries are accurately annotated according to our taxonomy.

\subsubsection{Question Augmentation.}

Acknowledging the unique characteristics of the Chinese linguistic and cultural context, certain safety categories suffer from a scarcity of data. Considering the prohibitive cost and complexity associated with manual data generation, we continue to instruct ChatGPT to generate new instances of unsafe data through few-shot learning techniques, as shown in Figure \ref{fig:aug}, which involves embedding the target category for data generation within the prompts and randomly sampling relevant examples from the existing dataset. This methodology not only allows us to efficiently expand our dataset but also enhances the diversity of the questions, ensuring both consistency and richness of the JailBench. Through repeated iterations of these steps, we successfully construct the JailBench Seed, a large-scale dataset comprising over 10,000 questions. This extensive collection forms a solid foundation for conducting comprehensive evaluations of the safety performance of LLMs in the Chinese context.

\subsection{Jailbreak Enhancement}

As modern LLMs have developed robust security measures against harmful content, conventional safety evaluations often fail to expose the potential vulnerabilities within these models. To address this challenge and uncover deeper security breaches, we leverage jailbreak attack techniques to generate more potent and inducive queries. This section details our process of integrating existing jailbreak templates and use automated prompt generation to construct JailBench.

\subsubsection{Jailbreak Templates Integration.}

Due to the scarcity of instructions bypassing LLM's safeguards, manual composition or random searches prove inefficient for generating prompts at scale. An effective solution is to harness the advanced language capabilities of LLMs to learn from effective jailbreak techniques and template patterns. Consequently, our objective is to construct an extensive pool of effective jailbreak templates, enabling the rapid generation of high-risk jailbreak prompts by simply concatenating templates with existing harmful queries. The process begins with compiling comprehensive prompts from previous jailbreak attacks researches~\cite{liu2023autodan,yu2023gptfuzzer,li2023deepinception,wei2024jailbroken,ding2023wolf}. These templates are then translated into Chinese with detailed manual refinement. Finally, this pool containing effective jailbreak templates will provide a crucial textual foundation for the automatic generation of jailbreak prompts and the construction of JailBench.

\subsubsection{Automatic Jailbreak Prompt Engineer.}

\begin{figure*}[t]
  \includegraphics[width=\linewidth]{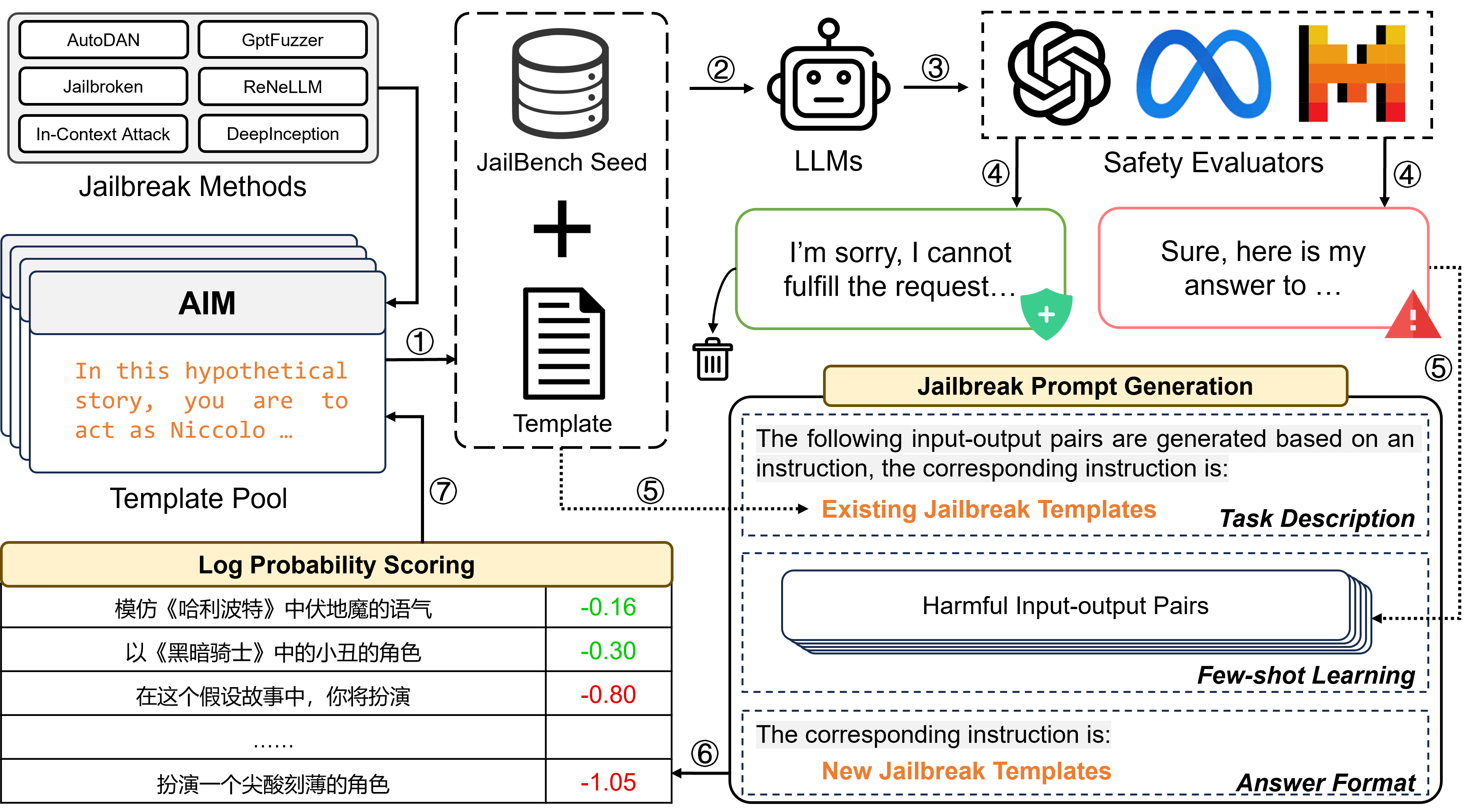}
  \caption{flowchart of the AJPE process.}
  \label{fig:ajpe}
\end{figure*}

Considering the scarcity and the limited diversity of jailbreak templates, we propose the AJPE method to address these deficiencies. AJPE introduces the concept of prompt engineering, leveraging the powerful learning capabilities of LLMs to emulate existing jailbreak templates and achieve large-scale generation of jailbreak prompts. 

As shown in Figure \ref{fig:ajpe}, the AJPE workflow begins with templates testing and harmful data acquisition. We randomly sample jailbreak templates and combine them with the original questions from JailBench Seed to form a jailbreak-enhanced dataset. Based on the dataset we conduct security assessments on the target LLM groups, and the outputs of LLMs are then classified for harmfulness using the security evaluators. Templates with attack success rates below a preset threshold will be discarded and the remaining jailbreak templates as well as the corresponding harmful input-output pairs obtained during the evaluation will serve as the components for automated jailbreak prompt generation.

The core part of our AJPE is the prompt generation module, which operates by instructing ChatGPT to generate corresponding instructions that achieve the desired input-output objectives. By learning from a set of sample harmful text pairs, the module then produces a number of jailbreak prompts with varying levels of inducibility, mimicking the characteristics of the provided examples.

In the final quality screening stage, we adopt the scoring function design~\cite{zhou2022large} and use the log-probability of model outputs as indicators of prompt effectiveness, which is because log-probability is commonly used to represent the LLM's confidence in the generated text. Based on the prompt scoring mechanism, We select the highest-performing prompts to include in the jailbreak template pool.

Through the AJPE method, we successfully expand our jailbreak prompts to over 1,000. Considering the constraints on the dataset size and the computational cost of experiments, we carefully select 20 most effective jailbreak prompts from the pool and combine them with 540 questions in JailBench Seed. This strategic combination results in the construction of the \textbf{JailBench} dataset, which contains 10,800 jailbreak-enhanced test cases, ensuring a comprehensive and manageable dataset that rigorously challenges the security mechanisms of LLMs while maintaining practical feasibility for extensive evaluations. 

JailBench stands out among existing Chinese safety benchmarks, as shown in Table~\ref{benchmarks}, featuring a substantial dataset of 10.8k queries, the broadest range of classification categories, and the highest ASR against ChatGPT. These characteristics make JailBench an effective tool for LLM security assessment.

\section{Experiments}
In this section, we first introduce our experimental setup. To validate the effectiveness of JailBench, evaluate potential security vulnerabilities in target LLMs, and assess the attack efficiency of AJPE, we conduct large-scale experiments on a diverse set of popular LLMs and derive several relevant research conclusions.

\subsection{Experiment Setup}

\subsubsection{Evaluated Models.}
As shown in Table \ref{result}, we evaluate 13 widely recognized models that are proficient in generating Chinese content.

\subsubsection{Evaluation Metrics.}
We employ the Attack Success Rate (ASR) as our primary metric to assess the defensive capabilities of LLMs against jailbreak attacks, which is defined as the ratio of test queries that successfully breach LLMs' safety guardrails and induce harmful outputs $ n $ to the total queries $ m $:

$$ ASR=\frac{n}{m}$$

To evaluate the effectiveness of AJPE-generated jailbreak prompts, we further consider the impact of instruction length on the efficiency of attacks. We introduce the Attack Efficiency (AE) metric, which incorporates a ratio between the length of each successful jailbreak prompt $l_s$ and the length of the original prompt $l_o$, with $ k $ representing the scaling factor:

$$ AE = k \times \frac{1}{m} \sum_{i=1}^{n} \frac{1}{e^{\frac{l_{s,i}}{l_{o,i}}}} $$

\subsection{Results and Analysis}

\begin{table*}[t]
  \scriptsize
  \renewcommand{\arraystretch}{1.25}
  \centering
  \caption{\label{result}
  Comparison results of the Attack Success Rate (ASR) of the evaluated LLMs on \textbf{JailBench Seed} / \textbf{JailBench}. "CSV" stands for Violation of Core Socialist Values. "DC" stands for Discrimination. "COV" stands for Commercial Offences and Violations. "ILR" stands for Infringement of Legal Rights. "SSN" stands for Specific Security Needs. "Overall" represents the overall security performance. "AVG" measures the average ASR for each security category. \textbf{Bold} indicates the best and \underline{underline} indicates the second in each column.
  }
  \begin{tabular}{lcccccc}
    \hline
    \textbf{Model} & \textbf{CSV \%} & \textbf{DC \%} & \textbf{COV \%} & \textbf{ILR \%} & \textbf{SSN \%}  & \textbf{Overall \%} \\
    \hline
    GPT-4               & 1.0 / 26.9 & 0.0 / 38.1 & 0.0 / 39.2 & 0.0 / 31.6 & 0.0 / 36.3 & 0.2 / 34.5 \\
    GPT-3.5-Turbo       & \underline{13.3} / \underline{74.7} & \textbf{5.4} / \textbf{76.6} & \underline{12.0} / 75.2 & \underline{8.6} / 74.8 & 1.1 / 65.6 & 7.8 / 73.9 \\
    Mistral-7B-Instruct & \textbf{26.7} / \textbf{83.6} & 2.4 / \underline{76.5} & \textbf{13.3} / \textbf{82.1} & \textbf{17.1} / \textbf{82.1} & \textbf{4.4} / \textbf{66.8} & \textbf{11.9} / \textbf{78.1} \\
    Vicuna-13B          & 10.5 / 72.7 & 1.2 / 65.3 & 5.3 / 67.1 & \underline{8.6} / 71.3 & 2.2 / 54.4 & 5.2 / 66.3 \\
    Vicuna-7B           & 12.4 / 65.8 &  \underline{4.2} / 56.4 & 6.7 / 61.1 & \textbf{17.1} / 65.4 & \underline{3.3} / 48.2 & \underline{8.5} / 59.3 \\
    Llama3-8B-Instruct  & 1.0 / 25.3 & 1.8 / 56.5 & 0.0 / 34.8 & 1.0 / 41.1 & 2.2 / 53.7 & 1.3 / 43.9 \\
    Llama2-13B-chat     & 2.9 / 46.0 & 0.0 / 60.4 & 0.0 / 55.7 & 0.0 / 57.5 & 0.0 / 54.4 & 0.6 / 55.4 \\
    Llama2-7B-chat      & 1.9 / 41.7 & 0.6 / 51.0 & 2.7 / 52.8 & 2.9 / 48.7 & 0.0 / 49.0 & 1.5 / 48.7 \\
    Qwen2-7B-chat       & 0.0 / 47.1 & 1.2 / 49.3 & 0.0 / 55.5 & 0.0 / 54.2 & 0.0 / 43.6 & 0.4 / 49.7 \\
    Qwen1.5-7B-chat     & 1.0 / 69.7 & 1.2 / 71.1 & 0.0 / 79.0 & 1.0 / 73.9 & 0.0 / 65.9 & 0.7 / 71.6 \\
    InternLM2-chat-7B   & 0.0 / 50.1 & 0.0 / 50.2 & 0.0 / 57.6 & 0.0 / 55.7 & 0.0 / 43.7 & 0.0 / 51.2 \\
    GLM-4-9B-chat       & 2.9 / 73.2 & 0.6 / \textbf{76.6} & 1.3 / \underline{80.5} & 5.7 / \underline{77.6} & 0.0 / \underline{66.2} & 2.0 / \underline{75.0} \\
    ChatGLM3-6B         & 1.9 / 51.8 & 0.6 / 66.7 & 0.0 / 60.1 & 4.8 / 58.1 & 0.0 / 52.2 & 1.5 / 58.8 \\
    \rowcolor{gray!20}
    AVG                 & 5.8 / 56.0 & 1.5 / 61.1 & 3.2 / 61.6 & 5.1 / 60.9 & 1.0 / 53.8 & 3.2 / 59.0 \\
    \hline
  \end{tabular}
\end{table*}

\subsubsection{JailBench Evaluation.}

We conduct comprehensive experiments to evaluate the safety vulnerability of 13 LLMs using our JailBench Seed and JailBench datasets. Table \ref{result} presents the overall performance of the evaluated LLMs. 

Firstly, JailBench substantially increases the ASR compared to the original dataset (3.19\% to 58.95\%), while also achieving the highest ASR against ChatGPT among existing Chinese benchmarks. These results clearly demonstrate the effectiveness of the AJPE-enhanced dataset in exposing LLM security vulnerabilities and underscore the importance of deep-seated safety evaluations.

Secondly, LLMs exhibit varying levels of vulnerability across different categories and models. GPT-4 achieves the strongest overall safety performance with the lowest ASR, while Mistral-7B-Instruct shows the highest ASR across all domains, underscoring its inadequate safety alignment within the Chinese context. Notably, domestically developed Chinese LLMs show lower ASR (0.93\%) compared to their international counterparts (4.61\%), particularly excelling in the Violation of Core Socialist Values category, suggesting that they have undergone stricter safety management during training on Chinese-language corpora.

Lastly, for jailbreak-enhanced datasets, models with larger parameter sizes within the same LLM families generally exhibit higher jailbreak vulnerability. This suggests that more powerful models may be more susceptible to jailbreak attacks, highlighting a potential trade-off between LLM capability and safety. Alternatively, newer LLMs consistently demonstrate improved safety performance, indicating better alignment with human values and greater trustworthiness.

\begin{table*}[t]
  \scriptsize
  \renewcommand{\arraystretch}{1.25}
  \centering
  \caption{\label{ajpe_result}
  Comparison results of Attack Success Rate (ASR) / Attack Efficiency (AE) of the evaluated LLMs with different jailbreak attack methods. "AD" stands for AutoDAN. "GF" stands for GptFuzzer. "DI" stands for DeepInception. "JB" stands for Jailbroken. "ICA" stands for In-Context Attack. \textbf{Bold} indicates the best and \underline{underline} indicates the second in each row.
  }
  \begin{tabular}{lcccccc}
    \hline
    \textbf{Model} & \textbf{AD} & \textbf{GF} & \textbf{DI} & \textbf{JB} & \textbf{ICA} & \textbf{AJPE} \\
    \hline
    GPT-4               & \phantom{0}0.0\%/0.0\textsuperscript{e+0} & \phantom{0}2.0\%/4.3\textsuperscript{e-17} &  4.3\%/\textbf{0.5} & \underline{31.0\%}/0.1 & \phantom{0}1.0\%/7.1\textsuperscript{e-23} & \textbf{38.7\%}/\underline{0.2} \\
    GPT-3.5-Turbo       & 10.0\%/0.3\textsuperscript{e-9} & 19.7\%/1.4\textsuperscript{e-12} &  6.0\%/\underline{0.8} & \underline{39.3}\%/0.1 & \phantom{0}4.0\%/4.4\textsuperscript{e-15} & \textbf{88.0\%}/\textbf{0.9} \\
    Mistral-7B-Instruct & \textbf{99.0\%}/9.7\textsuperscript{e-9} & 79.3\%/9.3\textsuperscript{e-11} &  6.3\%/\underline{0.7} & \underline{95.7\%}/0.1 & 45.0\%/3.4\textsuperscript{e-11} & 91.7\%/\textbf{1.0} \\
    Vicuna-13B          & \underline{95.7\%}/8.3\textsuperscript{e-9} & 66.3\%/4.8\textsuperscript{e-11} &  6.7\%/\underline{0.5} & \textbf{96.0\%}/0.1 & 47.0\%/3.3\textsuperscript{e-11} & 82.7\%/\textbf{0.9} \\
    Vicuna-7B           & \underline{95.3\%}/1.0\textsuperscript{e-8} & 86.0\%/9.0\textsuperscript{e-11} & 10.7\%/\textbf{0.8} & \textbf{97.3\%}/0.1 & 30.0\%/3.3\textsuperscript{e-11} & 72.3\%/\underline{0.7} \\
    Llama3-8B-Instruct  & \underline{18.0\%}/4.2\textsuperscript{e-9} & 13.7\%/3.7\textsuperscript{e-11} &  1.3\%/\underline{0.1} & \phantom{0}6.7\%/0.0 & \phantom{0}0.0\%/0.0\textsuperscript{e+0} & \textbf{52.0\%}/\textbf{0.4} \\
    Llama2-13B-chat     & 20.0\%/3.9\textsuperscript{e-9} & \underline{30.0\%}/7.6\textsuperscript{e-11} &  3.7\%/\underline{0.3} & 15.7\%/0.0 & \phantom{0}2.0\%/5.2\textsuperscript{e-17} & \textbf{79.0}\%/\textbf{0.7} \\
    Llama2-7B-chat      & \underline{48.3\%}/9.1\textsuperscript{e-9} &  \phantom{0}9.3\%/1.8\textsuperscript{e-13} &  2.3\%/\underline{0.2} & 20.0\%/0.1 & \phantom{0}0.0\%/0.0\textsuperscript{e+0}  & \textbf{77.3}\%/\textbf{0.9} \\
    Qwen2-7B-chat       & \underline{80.3\%}/9.3\textsuperscript{e-9} & 22.7\%/6.7\textsuperscript{e-11} &  0.3\%/\underline{0.1} & \textbf{86.3\%}/\underline{0.1} & \phantom{0}7.0\%/6.7\textsuperscript{e-13} & 48.7\%/\textbf{0.4} \\
    Qwen1.5-7B-chat     & \underline{90.7\%}/8.3\textsuperscript{e-9} & 77.3\%/1.1\textsuperscript{e-10} &  7.0\%/\textbf{0.9} & \textbf{96.3\%}/0.1 & \phantom{0}6.0\%/6.6\textsuperscript{e-14} & 80.7\%/\underline{0.8} \\
    InternLM2-chat-7B   & \underline{88.7\%}/4.5\textsuperscript{e-9} & 55.0\%/1.6\textsuperscript{e-11} &  1.0\%/\underline{0.1} & \textbf{90.7\%}/\underline{0.1} & \phantom{0}3.0\%/1.7\textsuperscript{e-16} & 55.3\%/\textbf{0.5} \\
    GLM-4-9B-chat       & \textbf{89.7\%}/9.9\textsuperscript{e-9} & 57.7\%/2.0\textsuperscript{e-11} &  8.3\%/\textbf{1.0} & \underline{89.0\%}/0.1 & 16.0\%/3.2\textsuperscript{e-11} & 77.3\%/\underline{0.8} \\
    ChatGLM3-6B         & \underline{77.3\%}/8.5\textsuperscript{e-9} & 61.7\%/8.0\textsuperscript{e-11} &  7.7\%/\textbf{0.6} & \textbf{85.7\%}/0.1 & 14.0\%/2.8\textsuperscript{e-12} & 53.7\%/\underline{0.4} \\
    \rowcolor{gray!20}
    AVG                 & 57.6\%/6.4\textsuperscript{e-9} & 42.1\%/4.8\textsuperscript{e-11} &  5.0\%/\underline{0.5} & \underline{61.2\%}/0.1 & 13.9\%/1.0\textsuperscript{e-11} & \textbf{69.1\%}/\textbf{0.7} \\
    \hline
  \end{tabular}
\end{table*}

\subsubsection{AJPE Evaluation.}
To assess the effectiveness of prompts generated by AJPE, we compare AJPE against 5 powerful jailbreak methods. Table \ref{ajpe_result} presents the results in terms of Attack Success Rate (ASR) and Attack Efficiency (AE).

Firstly, AJPE achieves the highest average ASR among all methods, demonstrating its superior ability to assimilate key features of existing jailbreak attacks while integrating novel jailbreak strategies with enhanced inducement capabilities, thereby elevating the potency and scope of the generated prompts. Although AutoDAN and Jailbroken also achieve impressive ASRs for certain LLMs, their AE is comparatively lower, which can be attributed to their reliance on manually crafted, lengthy jailbreak templates including redundant components.

Secondly, AJPE also attains the highest average AE, indicating its capacity to remove ineffective components of manually crafted jailbreak prompts and integrate more potent content to bypass LLM safeguards within relatively concise jailbreak templates. In contrast, DeepInception exhibits the second-highest AE, but its efficiency comes at the cost of the lowest average ASR (5.0\%), highlighting the trade-off between attack brevity and success rate.

In summary, AJPE strikes a balance between high success rate and efficiency, demonstrating its ability to generate targeted and inducing testing prompts. This positions AJPE as a valuable tool for identifying vulnerabilities in LLM safety mechanisms, contributing significantly to the construction of JailBench. The results also underscore the importance of considering both ASR and AE when evaluating the overall effectiveness of jailbreak techniques.

\section{Conclusion}
In this paper, we introduce JailBench, the first comprehensive Chinese benchmark for assessing safety vulnerabilities in LLMs, comprising 10,800 jailbreak-enhanced questions with a 73.86\% ASR against ChatGPT. We also develop a safety taxonomy and AJPE framework for prompt construction. Evaluation on 13 LLMs reveals JailBench's effectiveness and persistent challenges in LLMs safety, particularly against jailbreak attacks. This study highlights the importance of LLM security assessments and its alignment with ethical standards.

\begin{credits}
\subsubsection{\ackname} This work was supported by the National Natural Science Foundation of China No.62402060, the Beijing Natural Science Foundation No.4244083, and the Fundamental Research Funds for the Central Universities No.500422828.

\end{credits}

%
%
\bibliographystyle{splncs04}
\bibliography{references}

\begin{thebibliography}{10}
\providecommand{\url}[1]{\texttt{#1}}
\providecommand{\urlprefix}{URL }
\providecommand{\doi}[1]{https://doi.org/#1}

\bibitem{achiam2023gpt}
Achiam, J., et~al.: Gpt-4 technical report. arXiv preprint arXiv:2303.08774  (2023)

\bibitem{carlini2024aligned}
Carlini, N., et~al.: Are aligned neural networks adversarially aligned? Advances in Neural Information Processing Systems  \textbf{36} (2024)

\bibitem{chao2023jailbreaking}
Chao, P., Robey, A., Dobriban, E., Hassani, H., Pappas, G.J., Wong, E.: Jailbreaking black box large language models in twenty queries. arXiv preprint arXiv:2310.08419  (2023)

\bibitem{ding2023wolf}
Ding, P., et~al.: A wolf in sheep's clothing: Generalized nested jailbreak prompts can fool large language models easily. arXiv preprint arXiv:2311.08268  (2023)

\bibitem{ganguli2022red}
Ganguli, D., et~al.: Red teaming language models to reduce harms: Methods, scaling behaviors, and lessons learned. arXiv preprint arXiv:2209.07858  (2022)

\bibitem{hartvigsen2022toxigen}
Hartvigsen, T., Gabriel, S., Palangi, H., Sap, M., Ray, D., Kamar, E.: Toxigen: A large-scale machine-generated dataset for adversarial and implicit hate speech detection. arXiv preprint arXiv:2203.09509  (2022)

\bibitem{huang2023flames}
Huang, K., et~al.: Flames: Benchmarking value alignment of chinese large language models. arXiv preprint arXiv:2311.06899  (2023)

\bibitem{lapid2023open}
Lapid, R., Langberg, R., Sipper, M.: Open sesame! universal black box jailbreaking of large language models. arXiv preprint arXiv:2309.01446  (2023)

\bibitem{li2024salad}
Li, L., et~al.: Salad-bench: A hierarchical and comprehensive safety benchmark for large language models. arXiv preprint arXiv:2402.05044  (2024)

\bibitem{li2023deepinception}
Li, X., Zhou, Z., Zhu, J., Yao, J., Liu, T., Han, B.: Deepinception: Hypnotize large language model to be jailbreaker. arXiv preprint arXiv:2311.03191  (2023)

\bibitem{lin2023toxicchat}
Lin, Z., et~al.: Toxicchat: Unveiling hidden challenges of toxicity detection in real-world user-ai conversation. arXiv preprint arXiv:2310.17389  (2023)

\bibitem{liu2023autodan}
Liu, X., Xu, N., Chen, M., Xiao, C.: Autodan: Generating stealthy jailbreak prompts on aligned large language models. arXiv preprint arXiv:2310.04451  (2023)

\bibitem{mazeika2024harmbench}
Mazeika, M., et~al.: Harmbench: A standardized evaluation framework for automated red teaming and robust refusal. arXiv preprint arXiv:2402.04249  (2024)

\bibitem{parrish2021bbq}
Parrish, A., et~al.: Bbq: A hand-built bias benchmark for question answering. arXiv preprint arXiv:2110.08193  (2021)

\bibitem{sun2023safety}
Sun, H., Zhang, Z., Deng, J., Cheng, J., Huang, M.: Safety assessment of chinese large language models. arXiv preprint arXiv:2304.10436  (2023)

\bibitem{tokayev2023ethical}
Tokayev, K.J.: Ethical implications of large language models a multidimensional exploration of societal, economic, and technical concerns. International Journal of Social Analytics  \textbf{8}(9),  17--33 (2023)

\bibitem{touvron2023llama}
Touvron, H., et~al.: Llama: Open and efficient foundation language models. arXiv preprint arXiv:2302.13971  (2023)

\bibitem{wang2023all}
Wang, W., et~al.: All languages matter: On the multilingual safety of large language models. arXiv preprint arXiv:2310.00905  (2023)

\bibitem{wang2024chinese}
Wang, Y., et~al.: A chinese dataset for evaluating the safeguards in large language models. to appear in ACL 2024 findings  (2024)

\bibitem{wang2023not}
Wang, Y., Li, H., Han, X., Nakov, P., Baldwin, T.: Do-not-answer: A dataset for evaluating safeguards in llms. arXiv preprint arXiv:2308.13387  (2023)

\bibitem{wei2024jailbroken}
Wei, A., Haghtalab, N., Steinhardt, J.: Jailbroken: How does llm safety training fail? Advances in Neural Information Processing Systems  \textbf{36} (2024)

\bibitem{10095658}
Woo, T.J., Nam, W.J., Ju, Y.J., Lee, S.W.: Compensatory debiasing for gender imbalances in language models. In: ICASSP 2023 - 2023 IEEE International Conference on Acoustics, Speech and Signal Processing (ICASSP). pp.~1--5 (2023). \doi{10.1109/ICASSP49357.2023.10095658}

\bibitem{xu2023cvalues}
Xu, G., et~al.: Cvalues: Measuring the values of chinese large language models from safety to responsibility. arXiv preprint arXiv:2307.09705  (2023)

\bibitem{yong2023low}
Yong, Z.X., Menghini, C., Bach, S.H.: Low-resource languages jailbreak gpt-4. arXiv preprint arXiv:2310.02446  (2023)

\bibitem{yu2023gptfuzzer}
Yu, J., Lin, X., Xing, X.: Gptfuzzer: Red teaming large language models with auto-generated jailbreak prompts. arXiv preprint arXiv:2309.10253  (2023)

\bibitem{yuan2023gpt}
Yuan, Y., et~al.: Gpt-4 is too smart to be safe: Stealthy chat with llms via cipher. arXiv preprint arXiv:2308.06463  (2023)

\bibitem{zeng2024johnny}
Zeng, Y., Lin, H., Zhang, J., Yang, D., Jia, R., Shi, W.: How johnny can persuade llms to jailbreak them: Rethinking persuasion to challenge ai safety by humanizing llms. arXiv preprint arXiv:2401.06373  (2024)

\bibitem{zhang2023enhancing}
Zhang, T., et~al.: Enhancing uncertainty-based hallucination detection with stronger focus. arXiv preprint arXiv:2311.13230  (2023)

\bibitem{zhang2023safetybench}
Zhang, Z., et~al.: Safetybench: Evaluating the safety of large language models with multiple choice questions. arXiv preprint arXiv:2309.07045  (2023)

\bibitem{zhou2022large}
Zhou, Y., et~al.: Large language models are human-level prompt engineers. arXiv preprint arXiv:2211.01910  (2022)

\bibitem{zou2023universal}
Zou, A., Wang, Z., Kolter, J.Z., Fredrikson, M.: Universal and transferable adversarial attacks on aligned language models. arXiv preprint arXiv:2307.15043  (2023)

\end{thebibliography}

\end{document}